\documentclass[runningheads]{llncs}
\usepackage[T1]{fontenc}
\usepackage{amssymb}
\usepackage{graphicx}
\usepackage{tabularx}
\usepackage{multirow}
\usepackage{color}

\usepackage{hyperref}
\usepackage{booktabs}
\usepackage{bbding}
\usepackage{multicol}
\usepackage{amssymb}
\usepackage{amsmath}
\usepackage{lipsum}
\usepackage{bbm}

\begin{document}
\title{URL: Combating Label Noise for Lung Nodule Malignancy Grading}
\author{
Xianze Ai\inst{1,2,}\thanks{Equal contribution.} \and
Zehui Liao\inst{1,2,*} \and
Yong Xia\inst{1,2(}\Envelope\inst{)} 
}
\authorrunning{X. Ai et al.}
\institute{
Ningbo Institute of Northwestern Polytechnical University, Ningbo 315048, China\\
\and
National Engineering Laboratory for Integrated Aero-Space-Ground-Ocean Big Data Application Technology, School of Computer Science and Engineering, Northwestern Polytechnical University, Xi’an 710072, China \\
\email{yxia@nwpu.edu.cn}
}
\maketitle              

\begin{abstract}
Due to the complexity of annotation and inter-annotator variability, most lung nodule malignancy grading datasets contain label noise, which inevitably degrades the performance and generalizability of models.
Although researchers adopt the label-noise-robust methods to handle label noise for lung nodule malignancy grading, they do not consider the inherent ordinal relation among classes of this task.
To model the ordinal relation among classes to facilitate tackling label noise in this task, we propose a \textbf{U}nimodal-\textbf{R}egularized \textbf{L}abel-noise-tolerant (URL) framework.
Our URL contains two stages, the \textbf{S}upervised \textbf{C}ontrastive \textbf{L}earning (SCL) stage and the \textbf{M}emory pseudo-labels generation and \textbf{U}nimodal regularization (MU) stage.
In the SCL stage, we select reliable samples and adopt supervised contrastive learning to learn better representations.
In the MU stage, we split samples with multiple annotations into multiple samples with a single annotation and shuffle them into different batches. 
To handle label noise, pseudo-labels are generated using the similarity between each sample and the central feature of each class, and temporal ensembling is used to obtain memory pseudo-labels that supervise the model training.
To model the ordinal relation, we introduce unimodal regularization to keep the ordinal relation among classes in the predictions.
Moreover, each lung nodule is characterized by three orthographic views.
Experiments conducted on the LIDC-IDRI dataset indicate the superiority of our URL over other competing methods. 
Code is available at \href{https://github.com/axz520/URL}{https://github.com/axz520/URL}.
\end{abstract}

\section{Introduction}
Deep convolutional neural networks (DCNNs) have achieved impressive performance in lung nodule malignancy grading ~\cite{liu2019multi,chen2021artificial,gu2021performance} using chest computed tomography (CT). 
Their success depends on a large amount of reliably-labeled training data.
Medical professionals perform the annotation of chest CT scans on a slice-by-slice basis, which always requires a high degree of expertise and concentration and is labor-expensive and time-consuming~\cite{joskowicz2019inter}.
Due to the complexity of annotation and inter-annotator variability, the collected training data often contain label noise~\cite{liao2022learning}, which inevitably impair the performance and generalizability of the model trained with them.
Therefore, improving the robustness of the model against label noise is a crucial task for accurate and reliable lung nodule malignancy grading. 
In the broad area of pattern recognition, increasing research efforts have been denoted to the label noise issue, resulting in several innovative solutions.
Among them, some aim to identify noisy samples and reduce their impact by using semi-supervised algorithms, sample reweighting, or assigning them pseudo labels~\cite{wang2021proselflc,li2020dividemix,karim2022cnll}, while others aim to resist the label noise via designing a noise-robust loss function or estimating the noise transition matrix~\cite{sun2022pnp,liu2020early,wang2021learning}.
Despite their success in natural image processing, these solutions rarely consider the cases where each sample has several inconsistent annotations provided by different annotators, which is common in clinical diagnosis~\cite{liao2022learning,ju2022improving,liao2021modeling,liao2023transformer}.

Recently, a few methods have been proposed to deal with the label noise issue in medical image classification tasks, where each sample may have one unreliable or more inconsistent annotations~\cite{lei2022meta,ju2022improving,liao2022learning}.
An intuitive solution is to generate proxy labels, such as using the average / median / max voting of multiple annotations~\cite{lei2022meta,wu2019learning,xu2020mscs}. 
Besides, Jensen et al.~\cite{jensen2019improving} introduced a label sampling strategy that randomly selects the proxy label from the multiple annotations of each sample. 
Ju et al.~\cite{ju2022improving} proposed an uncertainty estimation-based framework that selects reliable samples using uncertainty scores and proceeds with course learning. 
Liao et al.~\cite{liao2022learning} proposed a ‘divide-and-rule’ model which reduces the impact of samples with inconsistent and unreliable labels by introducing the attention mechanism.
Although these methods can tackle the label noise in the scenario of multiple annotations, they do not take into account the inherent ordinal relation among classes in grading tasks.
The probabilities of neighboring labels should decrease with the increase of distance away from the ground truth. For instance, a lung nodule with a ground-truth malignancy of 2 is more likely to be misclassified into the categories of malignancy 1 and 3, instead of the categories of malignancy 4 and 5. In other words, the distribution of class transition probabilities is unimodal.
It should be noted that, although the ordinal relation among classes has been studied using the random forest, meta-learning, and ordinal regression in previous research on lung nodule malignancy grading ~\cite{lei2022meta,wu2019learning}, such research ignores the label noise issue.

In this paper, we propose a \textbf{U}nimodal-\textbf{R}egularized \textbf{L}abel-noise-tolerant (URL) framework for lung nodule malignancy grading using chest CT.
Under the URL framework, a nodule malignancy grading model can be trained in two steps, including warming up on reliable samples and fine-tuning on noisy samples.
In the warming-up step, reliable samples are first selected by adopting the negative learning strategy~\cite{kim2019nlnl} and then employed to warm up the grading model using contrastive learning.
In the fine-tuning step, two tricks are designed to alleviate the impact of noisy labels.
First, the memory pseudo-label generation (MPLG) module is constructed to generate pseudo-labels according to the feature similarity between each sample and the mean feature of each class and to improve those pseudo-labels by applying the temporal ensembling technique to them.
Second, considering the fact that the ordinal relation among classes means that the probability of each class label follows a unimodal distribution, we apply a unimodal regularization to the predicted probabilistic labels of each sample, forcing the distribution to have a single mode.
We have evaluated our URL framework against six competing methods on the LIDC-IDRI dataset~\cite{armato2011lung} and achieved state-of-the-art performance.

The main contributions of this work are as follows.
(1) We identify the importance of the inherent ordinal relation among classes in the task lung nodule malignancy grading with noisy labels and thus propose the URL framework to tackle this issue.
(2) Based on the ordinal relation among classes, we design a unimodal regularization to constrain the predicted probabilistic class label, leading to improve grading performance.
(3) Experimental results indicate that our RUL framework outperforms six competing methods in combating label noises for the lung nodule malignancy grading.

\section{Method}
\subsection{Problem Definition and Overview}
We define the dataset $D=\{(x_i,y_{i}^{1},y_{i}^{2},...,y_{i}^{s_i})\}_{i=1}^N$ for noisy $C$-class classification problem, 
where $x_i$ is the $i$-th image, $s_i$ is the number of annotations of $x_i$, $y_i^j$ is the $j$-th annotation of $x_i$, and $N$ is the number of samples in dataset $D$. Note that $y_{i}^{j} \in \{0, 1\}^{C}$ is the one-hot label over $C$ classes and it might be incorrect.
Our goal is to train a robust classification model using noisy training set $D$ for the lung nodule malignancy grading task.

\begin{figure*}[t!]  
 \includegraphics[width=\linewidth]{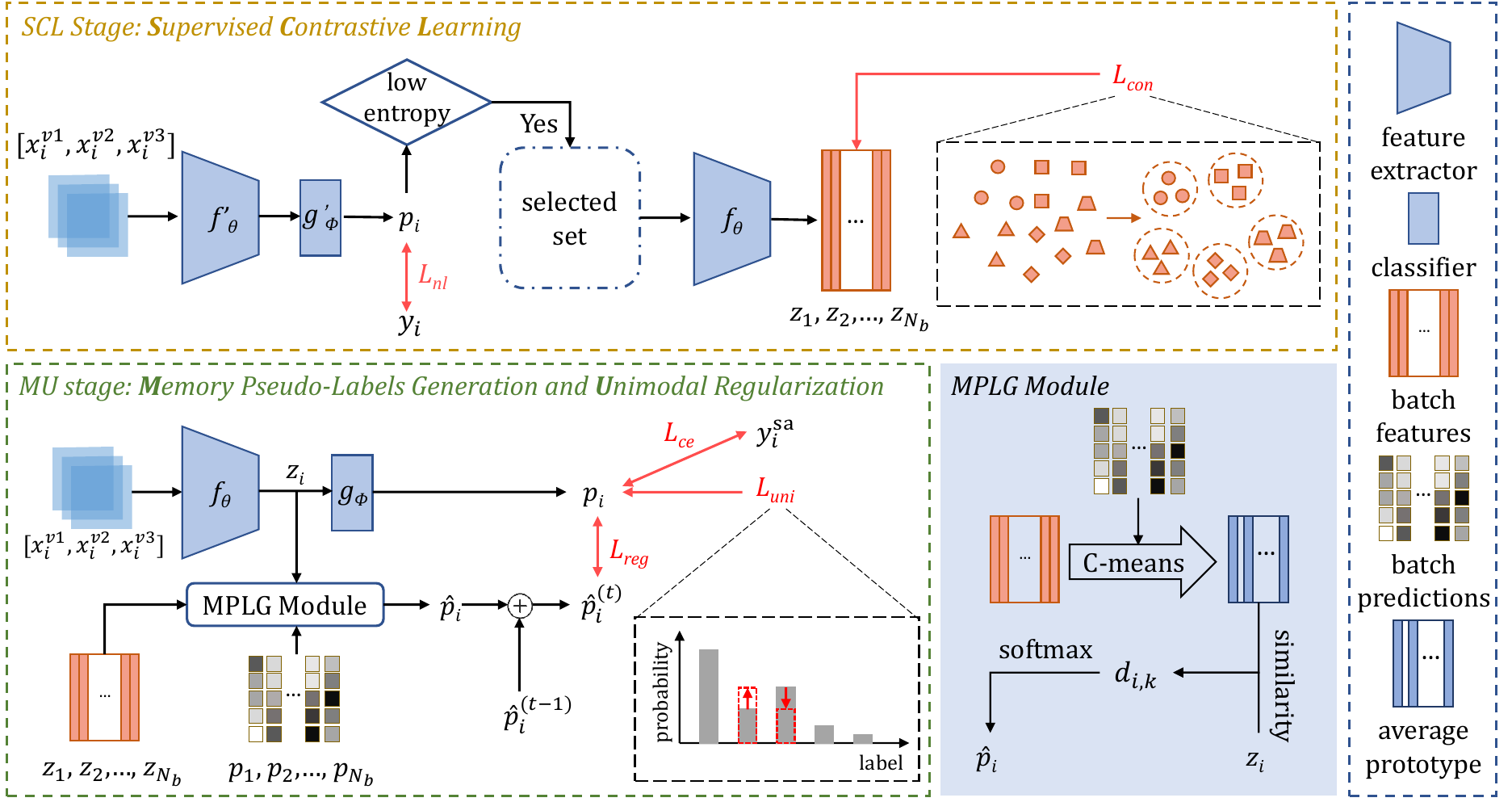}
 {\caption{\small The overview of our proposed URL framework. 
 [$x_i^{v1}, x_i^{v2}, x_i^{v3}$] means that three 2D views of the $i$-th image are concatenated at channel level, $z_i$ is the feature of $x_i$ and $y_i=y_{i}^1 \vee y_{i}^2 \vee ... \vee y_{i}^{s_i}$. $N_b$ is the batch size. $z_1,z_2,...z_{N_b}$ is batch features and $p_1,p_2,...p_{N_b}$ is batch predictions.
 $\hat{p}_{i}^{(t)}$ is the memory pseudo-label of $x_i$ at the $t$-th epoch.
 $y_i^{sa}$ is one of the candidate labels of $x_i$.
}  
 \label{fig:framework}
}
\end{figure*}

The proposed framework is shown in Fig~\ref{fig:framework}. There are two stages in our URL framework.
In the SCL stage, we first select reliable samples through negative learning and then train a feature extractor $f_\theta$ using these selected samples via supervised contrastive learning.
In the MU stage, pseudo-labels are generated to handle label noise using the MPLG module, and unimodal regularization is introduced to model the inter-class ordinal relation.

Note that taking into account balancing performance and memory consumption, we suggest using multi-view 2D slices instead of 3D images. 
Specifically, given an input image $x_i$, we extract three 2D slices on the axial, sagittal, and coronal planes ($i.e.$, $x_{i}^{v1}, x_{i}^{v2}, x_{i}^{v3}$), and concatenate them at the channel-wise as the input.
We now delve into the details of our framework.

\subsection{SCL Stage}
\noindent\textbf{Negative Learning for Reliable  Sample Selection.} There is inconsistency in multiple annotations, but their complementary labels are more reliable. 
Take $C=5$ for example, if three annotators individually conclude that a nodule malignancy is 3, 4, and 5, then we consider that the nodule is highly unlikely to be classified as 1 or 2. Hence, negative learning which uses complementary labels can ensure the reliability of annotations. The adopted backbone contains an encoder $f_\theta'$ and fully-connected layers $g_\phi'$ followed by a softmax layer $S$.
Given an input image $x_i$, its prediction is calculated as follows
\begin{equation}
\label{eq:p_i}
p_i = S(g_\phi'(f_\theta'(concat(x_{i}^{v1}, x_{i}^{v2}, x_{i}^{v3})))),
\end{equation}
where $p_i$ is the predicted probabilistic vector. The negative learning loss is calculated by

\begin{equation}
\label{eq:loss_nl}
\mathcal L_{nl} = - (\mathbf{1}-y_i) \log (\mathbf{1}-p_i),
\end{equation}
where $y_i=y_{i}^1 \vee y_{i}^2 \vee ... \vee y_{i}^{s_i}$ and $\mathbf{1}$ is all-one vector. 
We select the first $\frac{M}{C}$ low-entropy samples in each class as a reliable set $D_{r}=\{x_i, y_i'\}_{i=1}^{M}$, where $y_i'$ is the predicted label of the backbone.

\noindent\textbf{Supervised Contrastive Learning.} Supervised contrastive learning~\cite{khosla2020supervised} is powerful in representation learning, but it is degraded when there are noisy labels~\cite{li2022selective}. 
Therefore, we perform supervised contrastive learning using $D_{r}$ to learn better representations. 
It can maximize the feature similarities of the different classes. 
The feature of $x_i$ is calculated as $z_i=f_\theta(concat(x_{i}^{v1}, x_{i}^{v2}, x_{i}^{v3}))$, where $f_\theta$ is the other encoder.
In a mini-batch, we select samples with the same label as $x_i$ as positive examples and select samples with different labels from $x_i$ as negative examples. The supervised contrastive loss is calculated as follows: 
\begin{equation}
\label{eq:loss_con}
  \mathcal {L}_{con}
  =-\sum\limits_{i\in I}\frac{1}{|U(i)|}\sum\limits_{u\in U(i)}\log{\frac{\text{exp}\left({z}_i\cdot{z}_u/\tau\right)}{\sum\limits_{a\in A(i)}\text{exp}\left({z}_i\cdot{z}_a/\tau\right)}}.
\end{equation}
Here, $i \in I = \{1,2,...,N_b\}$ is the index of batch samples and $A(i) = I \setminus \{i\}$. $U(i) = \{u \in A(i) : y'_u = y'_i\}$ is the set of indices of positive examples and $|U(i)|$ is its cardinality, and $\tau$ is a temperature parameter.

\subsection{MU Stage}
Based on the encoder $f_\theta$ trained by supervised contrastive learning, we conduct a lung nodule malignancy grading model which contains an encoder $f_\theta$, fully-connected layers $g_\phi$ followed by a softmax layer $S$, and the MPLG module. 
We split training samples with multiple annotations into multiple samples with a single annotation and the reorganized training set is denoted as $D_{a}=\{x_i, y_i^{sa}\}_{i=1}^{N_{sa}}$. 
Given an input image $x_i$, the prediction is calculated as $p_i=S(g_\phi(f_\theta(concat(x_i^{v_1},x_i^{v_2},x_i^{v_3})))$ which is supervised by its label $y_i^{sa}$ using following cross-entropy Loss:  
\begin{equation}
\label{eq:loss_ce}
L_{ce}=- y_i^{sa} \log p_i.
\end{equation}

\noindent\textbf{Memory Pseudo-Labels Generation.} 
To handle label noise, we first generate pseudo-labels using MPLG Module. 
In a mini-batch, we calculate the central feature $\overline{z}_k$ of $k$-th class as follows
\begin{equation}
\label{eq:zk}
   \overline{z}_{k} = \frac{1}{N_{k}}\sum_{i=1}^{N_b} \mathbbm{1}\left[ \hat{y}_i=k \right]z_i,
\end{equation}
where $k \in \{1,2,...,C\}$, $\hat{y}_i=argmax(p_i)$ and $N_k$ is the number of samples that satisfy $\hat{y}_i=k$. $\mathbbm{1}[A]$ means the indicator of the event A. 
We calculate the feature similarity between the feature $z_i$ and $k$-th central feature $\overline{z}_k$ by the cosine distance as $d_{i,k} = \frac{z_i \overline{z}_k^T}{\| z_i \| \| \overline{z}_k \|}.$ Then we calculate pseudo-label $\hat{p}$ shown as follows:
\begin{equation}
\label{eqn:p_hat}
   \hat{p}_{i,k} ={\frac{\text{exp}\left(d_{i,k} / \tau\right)}{\sum_{j=1}^C\text{exp}\left(d_{i,j} / \tau\right)}}.
\end{equation}
Inspired by early learning and memorization phenomena~\cite{liu2020early}, we initialize the memory pseudo-label $\hat{p}_{i}^{(0)}$ with zeros and use temporal ensembling to update it as follows
\begin{equation} 
\label{eqn:p_hat_t}
\hat{p}_{i}^{(t)} = \beta \hat{p}_{i}^{(t-1)} + (1-\beta)\hat{p_i}.
\end{equation}
We maximize the inner product between the model output and the memory pseudo-label shown as follows:

\begin{equation} 
\mathcal{L}_\text{reg} = - \log(1-\langle \hat{p}^{(t)}_i,p_i \rangle).
\label{eq:loss_reg}
\end{equation}

\noindent\textbf{Unimodal Regularization.} 
In order to tackle label noise in the grading task according to its characteristics, we model the inter-class ordinal relation for facilitating label-noise-robust learning.
Hence, we introduce a unimodal regularization to constrain the class ordinal relation of the prediction $p_i$ as follows:

\begin{equation} 
\mathcal{L}_\text{uni} = \sum_{k=1}^{\hat{y}} max(0, p_{i,k}-p_{i,k+1}) + \sum_{k={\hat{y}}}^{C} max(0, p_{i,k+1}-p_{i,k}).
\label{eq:loss_ord}
\end{equation}

Finally, the objective loss function of the MU stage is denoted as follows:

\begin{equation} 
\mathcal L = L_{ce} + \alpha_1L_{reg} + \alpha_2L_{uni},
\label{eq: loss_ord}
\end{equation}
where $\alpha_1$ and $\alpha_2$ are hyper-parameters. 

\section{Experiments and Results}
\subsection{Dataset and Experimental Setup}

\noindent\textbf{Dataset.} We use the largest public lung nodule dataset LIDC-IDRI~\cite{armato2011lung} for this study. It contains 2568 lung nodules from 1018 chest CT scans. 
Each nodule is individually annotated by up to four annotators. 
The malignancy of each nodule is assessed using a rating scale ranging from 1 to 5, denoting an ascending malignancy. 
We split the dataset as shown in Table~\ref{tab:table1}. 
In the test set, all samples are annotated by multiple annotators and the annotations are consistent. 
20$\%$ of the training data is split as the validation set. 
We can approach the optimal model by approximately maximizing the accuracy on noisy distribution~\cite{chen2021robustness}. 
For each 3D image, we first sample each scan to a cubic voxel size of $1.0\times1.0\times1.0 mm^3$ and crop a $64\times64\times64$ cube which contains a lung nodule in its center. Before multi-view concatenation, all patches are resized to 224 $\times$ 224. For data augmentation, we employ random horizontal flipping and vertical flipping.

\begin{table}[t]
\caption{Malignancy distribution in the training/test set of the LIDR-IDRI dataset.}
\setlength{\tabcolsep}{10pt}
\label{tab:table1}
\footnotesize
\centering
\begin{tabular}{ccccccc}
\hline
\hline
\multirow{2}{*}{Datasets} & \multicolumn{5}{c}{Malignancy} & \multirow{2}{*}{Number} \\ \cline{2-6}
                          & 1    & 2    & 3    & 4   & 5   &                         \\ \hline
training set              & 400  & 1140   & 1476  & 708  & 370  & 2174                     \\ 
test set                  & 149  & 63   & 143  & 10  & 29  & 394                    \\ \hline \hline
\end{tabular}
\end{table}

\noindent \textbf{Implementation Details.} We use EfficientNet-B0~\cite{tan2019efficientnet} as the backbone that is pre-trained on the ImageNet dataset~\cite{deng2009imagenet}.
Adam~\cite{kingma2014adam} optimizer with a batch size of 32 is used to optimize the model. 
All competing methods are trained for 30 epochs with an initial learning rate of 0.0001 and we use the exponential decay strategy with a decay rate of 0.95. 
The best checkpoint used for reliable sample selection is obtained using the early stop strategy on the validation set during negative learning.
In the SCL stage, the encoder is pre-trained for 10 epochs. 
The experiments were performed on the PyTorch framework using a workstation with one NVIDIA GTX 1080Ti GPU.
The experimental results were reported over three random runs.
Hyper-parameters are set as $M=200$, $\beta = 0.9, \tau =  0.1, \alpha_1 = 0.8$ and $\alpha_2=3$.

\noindent \textbf{Evaluation Metrics.} 
We evaluate the performance of the 5-class classification problem (from 1 to 5) and the 3-class classification problem (benign, unsure, and malignant) in the lung nodule malignancy grading tasks.
\begin{table}[t]
\caption{Performance (mean $\pm$ standard deviation) of our URL framework and other competitors in the lung nodule malignancy grading task (\textbf{5-class classification}). The best and second-best results are highlighted in  $\textbf{bold}$/$\underline{underlined}$, respectively.}
\label{tab:table2}
\footnotesize
\centering
\setlength{\tabcolsep}{8pt}
\begin{tabular}{lccc}
\hline
\hline
 \multirow{2}{*}{Method} & \multicolumn{3}{c}{Results (\%)}  \\ \cline{2-4} 
  & Accuracy     & AUC          	& F1-score\\ \hline
AVE                              & 58.21 $\pm$ 0.56  & 78.55 $\pm$ 0.57  & 46.55 $\pm$ 0.52                        \\ 
LS~\cite{jensen2019improving}                   & 65.39 $\pm$ 0.40  & 83.35 $\pm$ 0.35   & 51.16 $\pm$ 0.34                      \\ \hline
UCL~\cite{li2022unimodal}               & 65.73 $\pm$ 0.39  & 84.35 $\pm$ 0.66 & 54.09 $\pm$ 0.22         \\ \hline
DU~\cite{ju2022improving}                     & 67.08 $\pm$ 0.26  & 82.56 $\pm$ 0.17   & $\underline{55.14}$ $\pm$ 0.69                        \\ \hline
SCE~\cite{wang2019symmetric}                     & 69.28 $\pm$ 0.56  & 83.37 $\pm$ 0.35 & 54.01 $\pm$ 0.41                       \\
ELR~\cite{liu2020early}                   & 70.16 $\pm$ 0.27  & 83.18 $\pm$ 0.13   & 53.95 $\pm$ 0.30                         \\
NCR~\cite{iscen2022learning}                   & $\underline{71.23}$ $\pm$ 0.18    & $\underline{85.17}$ $\pm$ 0.38   & 54.95 $\pm$ 0.42                        \\ \hline
Ours                             & $\textbf{73.18}$ $\pm$ 0.18   & $\textbf{85.82}$ $\pm$ 0.43  & $\textbf{57.25}$ $\pm$ 0.40                      \\ \hline \hline
\end{tabular}
\end{table}
\begin{table}[t]
\caption{Performance (mean $\pm$ standard deviation) of our URL framework and other competitors in the lung nodule malignancy grading task (\textbf{3-class classification}). The best and second-best results are highlighted in  $\textbf{bold}$/$\underline{underlined}$, respectively.}
\label{tab:table3}
\footnotesize
\centering
\setlength{\tabcolsep}{8pt}
\begin{tabular}{lccc}
\hline
\hline
\multirow{2}{*}{Method} & \multicolumn{3}{c}{Results (\%)}  \\ \cline{2-4} 
  & Accuracy     & AUC          	& F1-score\\ \hline
AVE                              & 67.25 $\pm$ 0.70  & 85.77 $\pm$  0.43  & 68.56 $\pm$ 0.48                        \\ 
LS~\cite{jensen2019improving}                   & 68.28 $\pm$ 0.28  & 88.75 $\pm$ 0.27   & 68.42 $\pm$ 0.60                         \\ \hline
UCL~\cite{li2022unimodal}               & 72.01 $\pm$ 0.24  & 88.87 $\pm$ 0.43 & 71.84 $\pm$ 0.15         \\ \hline
DU~\cite{ju2022improving}                     & 72.55 $\pm$ 0.14  & 87.54 $\pm$ 0.02   & 72.54 $\pm$ 0.75                        \\ \hline
SCE~\cite{wang2019symmetric}                     & 72.67 $\pm$ 0.40  & 87.93 $\pm$ 0.54 & 70.60 $\pm$ 0.41                        \\
ELR~\cite{liu2020early}                   & 74.02 $\pm$ 0.27  & 89.70 $\pm$ 0.10   & 72.06 $\pm$ 0.61                         \\
NCR~\cite{iscen2022learning}                   & $\underline{76.30}$ $\pm$ 0.10     & \underline{91.22} $\pm$ 0.02   & $\underline{76.70}$ $\pm$ 0.37                       \\ \hline
Ours                             & $\textbf{77.15}$ $\pm$ 0.17   & $\textbf{91.58}$ $\pm$ 0.21  & $\textbf{77.41}$ $\pm$ 0.38                       \\ \hline \hline
\end{tabular}
\end{table}
And accuracy, F1-score, and area under the ROC curve (AUC) are used as the metrics.

\subsection{Comparative Experiments}
We compared our URL framework with seven methods, including 
(1) two baseline methods: Average (AVE) uses the average proxy label, while Label Sampling (LS)~\cite{jensen2019improving} randomly selects one proxy label from multiple annotations;
(2) one method for modeling ordinal relation: Unimodal-Concentrated Loss (UCL)~\cite{li2022unimodal} combines concentrated loss and unimodal loss for ordinal classification;
(3) three methods for tackling noisy data with single annotation: SCE~\cite{wang2019symmetric}, ELR~\cite{liu2020early}, and NCR~\cite{iscen2022learning};  
(4) and one method for handling noisy data with several annotations: Dual Uncertainty (DU)~\cite{ju2022improving} evaluates the uncertainty of each sample and then adopts sample reweighting and curriculum training. 
Multi-view concatenation is adopted for all competing methods.
Table~\ref{tab:table2} shows the results of the 5-class classification task and Table~\ref{tab:table3} shows the results of the 3-class classification task. 
Experimental results demonstrate that our URL framework achieves the highest accuracy, AUC, and F1 score.

\subsection{Ablation Analysis}
We conducted ablation studies on the LIDC-IDRI dataset to investigate the effectiveness of each component of our URL, respectively.
Table~\ref{tab:table4} shows the results of the 5-class classification task and Table~\ref{tab:table5} shows the results of the 3-class classification task. 
Compared to taking a single view as input, taking multiple views as input provides more information and performs better.
And then we investigate the contribution of $L_{uni}$, $L_{reg}$, and $L_{con}$. 
Experimental results reveal that the performance of our URL framework is degraded more or less when $L_{uni}$, $L_{reg}$, or $L_{con}$ is removed.

\begin{table}[t]
\caption{
Performance (mean $\pm$ standard deviation) of our URL framework and its four variants in the lung nodule malignancy grading task (\textbf{5-class classification}). The best results are highlighted in  $\textbf{bold}$. `SV' and `MV' mean single-view and multi-view, respectively.
}
\label{tab:table4}
\footnotesize
\centering
\setlength{\tabcolsep}{6pt}
\begin{tabular}{lccc}
\hline
\hline
\multirow{2}{*}{method} & \multicolumn{3}{c}{Results (\%)}  \\ \cline{2-4}
& Accuracy     & AUC           	& F1-score\\ \hline
Ours    & $\textbf{73.18}$ $\pm$ 0.18   & $\textbf{85.82}$ $\pm$ 0.43  & $\textbf{57.25}$ $\pm$ 0.40 \\ \hline
MV+Baseline+$L_{con}$+$L_{reg}$  & 71.40 $\pm$ 0.02  & 85.40 $\pm$ 0.31  & 56.04 $\pm$ 0.19 \\
MV+Baseline+$L_{con}$    & 68.02 $\pm$ 0.55 & 83.70 $\pm$ 0.13 & 53.32 $\pm$ 0.02 \\
MV+Baseline & 65.93 $\pm$ 0.69  & 83.32 $\pm$ 0.39  & 52.59 $\pm$ 0.52 \\
SV+Baseline & 61.57 $\pm$ 0.67  & 81.30 $\pm$ 0.82  & 48.33 $\pm$ 0.31             \\ \hline \hline
\end{tabular}
\end{table}

\begin{table}[t]
\caption{Performance (mean $\pm$ standard deviation) of our URL framework and its four variants in the lung nodule malignancy grading (\textbf{3-class classification}). The best results are highlighted in  $\textbf{bold}$. `SV' and `MV' mean single-view and multi-view, respectively.}
\label{tab:table5}
\footnotesize
\centering
\setlength{\tabcolsep}{6pt}
\begin{tabular}{lccc}
\hline
\hline
\multirow{2}{*}{method} & \multicolumn{3}{c}{Results (\%)}  \\ \cline{2-4}
& Accuracy     & AUC           	& F1-score\\ \hline
Ours   & $\textbf{77.15}$ $\pm$ 0.17   & $\textbf{91.58}$ $\pm$ 0.21  & $\textbf{77.41}$ $\pm$ 0.38 \\ \hline
MV+Baseline+$L_{con}$+$L_{reg}$ & 75.63 $\pm$ 0.55  & 91.24 $\pm$ 0.17  & 75.88 $\pm$ 0.23 \\
MV+Baseline+$L_{con}$ & 71.73 $\pm$ 0.53 & 90.32 $\pm$ 0.70  & 72.57 $\pm$ 0.32 \\  
MV+Baseline   & 68.74 $\pm$ 0.49   & 88.65 $\pm$ 0.50  & 68.63 $\pm$ 0.44 \\
SV+Baseline   & 66.41 $\pm$ 0.22  & 85.77 $\pm$ 0.48  & 64.92 $\pm$ 0.78            \\ \hline \hline
\end{tabular}
\end{table}

\section{Conclusion}
In this paper, we propose to model the inter-class ordinal relation for facilitating the label-noise-robust learning for the lung nodule malignancy grading task.
To achieve this, we propose the URL framework. 
We generate memory pseudo labels by calculating feature similarity to handle label noise and introduce unimodal regularization to model the inter-class ordinal relation. 
Moreover, supervised contrastive learning is used to learn better representations. 
We conducted experiments on the LIDC-IDRI dataset and the results show that our URL framework performs better than other competing methods significantly.
Ablation studies demonstrate the contribution of modeling the class ordinal relation to label noise learning.

\subsubsection{Acknowledgement.} 
This work was supported 
in part by the Ningbo Clinical Research Center for Medical Imaging under Grant 2021L003 (Open Project 2022LYKFZD06),
in part by the Natural Science Foundation of Ningbo City, China, under Grant 2021J052,
in part by the National Natural Science Foundation of China under Grant 62171377, 
in part by the Key Technologies Research and Development Program under Grant 2022YFC2009903 / 2022YFC2009900,
and in part by the Innovation Foundation for Doctor Dissertation of Northwestern Polytechnical University under Grant CX2022056.

\bibliographystyle{splncs04}
\bibliography{mybib}
\end{document}